# Support Vector classifiers for Land Cover Classification

**Mahesh Pal**
Lecturer, department of Civil engineering
National Institute of Technology
Kurukshetra 136119
Haryana (India)
mpce_pal@yahoo.co.uk

**Paul M. Mather**
Prof., School of geography
University of Nottingham
University Park
Nottingham, NG7 2RD, UK
paul.mather@nottingham.ac.uk

## 1. Introduction

Much research effort in the past ten years has been devoted to analysis of the performance of artificial neural networks in image classification (Benediktsson et al., 1990; Heermann and Khazenie, 1992). The preferred algorithm is feed-forward multi-layer perceptron using back-propagation, due to its ability to handle any kind of numerical data, and to its freedom from distributional assumptions. Although neural networks may generally be used to classify data at least as accurately as statistical classification approaches a number of studies have reported that users of neural classifiers have problems in setting the choice of various parameters during training (Wilkinson, 1997). The choice of architecture of the network, the sample size for training, learning algorithms, and number of iterations required for training are some of these problems. A new classification system based on statistical learning theory (Vapnik, 1995), called the support vector machine has recently been applied to the problem of remote sensing data classification (Huang et al., 2002; Zhu and Blumberg, 2002; Gualtieri and Cromp, 1998). This technique is said to be independent of the dimensionality of feature space as the main idea behind this classification technique is to separate the classes with a surface that maximise the margin between them, using boundary pixels to create the decision surface. The data points that are closest to the hyperplane are termed "support vectors". The number of support vectors is thus small as they are points close to the class boundaries (Vapnik, 1995). One major advantage of support vector classifiers is the use of quadratic programming, which provides global minima only. The absence of local minima is a significant difference from the neural network classifiers.

## 2. Classification Methods

Three classification algorithms used for this study are the maximum-likelihood, multi-layer backpropagation neural network and support vector classifier. A brief summary of these classifiers is given below.

### 2.1 Maximum-Likelihood Classifier

The Maximum likelihood Classifier (MLC) is based on the assumption that the members of each class are normally distributed in feature space. MLC is a pixel-based method and can be defined as follows: a pixel with an associated observed feature vector X is assigned to class $c_j$ if

$$X \in c_j \text{ if } g_j(X) > g_k(X) \text{ for all } j \neq k,\ j, k = 1, \ldots\ldots, N$$

For multivariate Gaussian distributions $g_k(X)$ is given by:

$$g_k(X) = \ln(p(c_j)) - \frac{1}{2}\ln\left|\sum_k\right| - \frac{1}{2}(X - M_k)^t \sum_k^{-1}(X - M_k)$$

where $M_k$ and $\sum_k$ are the sample mean vector and covariance matrix of class k, and $g_k$ is the discriminating function.

Implementation of the MLC involves the estimation of class mean vectors and covariance matrices using training patterns chosen from known examples of each particular class.

### 2.2 Artificial Neural Network Classifier

A feed-forward artificial neural network (ANN) is used in this study. This is the most widely used neural network model, and its design consists of one input layer, at least one hidden layer, and one output layer. Each layer is made up of non-linear processing units called neurons, and the connections between neurons in successive layers carry associated weights. Connections are directed and allowed only in the forward direction, e.g. from input to hidden, or from hidden layer to a subsequent hidden or output layer. Non-linear processing is performed by applying an activation function to the summed inputs to a unit. Back-propagation is a gradient-descent algorithm that minimises the error between the output of the training input/output pairs and the actual network outputs (Bishop, 1995). Therefore, a set of input/output pairs is repeatedly presented to the network and the error is propagated from the output back to the input layer. The weights on the backwards path through the network are updated according to an update rule and a learning rate. ANNs are not solely specified by the characteristics of their processing units and the selected training or learning rule. The network topology, i.e. the number of hidden layers, the number of units, and their interconnections, also has an influence on classifier performance. In this study we use the network architecture and number of patterns used for training suggested by Kavzoglu (2001).

### 2.3 Support vector classifier

In the two-class case, a support vector classifier attempts to locate a hyperplane that maximises the distance from the members of each class to the optimal hyperplane. The principle of a support vector classifier is briefly described next.

Assume that the training data with *k* number of samples is represented by $\{\mathbf{x_i}, y_i\}$, i = 1, …, k, where $\mathbf{x} \in \mathbf{R}^n$ is an n-dimensional vector and $y \in \{-1,+1\}$ is the class label. These training patterns are said to be linearly separable if a vector **w** (which determining the orientation of a discriminating plane) and a scalar *b* (determine offset of the discriminating plane from origin) can be defined so that inequalities (1) and (2) are satisfied.

$$\mathbf{w} \cdot \mathbf{x_i} + b \geq +1 \qquad \text{for all } y = +1 \tag{1}$$

$$\mathbf{w} \cdot \mathbf{x_i} + b \leq -1 \qquad \text{for all } y = -1 \tag{2}$$

The aim is to find a hyperplane which divides the data so that that all the points with the same label lie on the same side of the hyperplane. This amounts to finding **w** and *b* so that

$$y_i \left(\mathbf{w} \cdot \mathbf{x}_i + b\right) > 0 \tag{3}$$

If a hyperplane exists that satisfies (3), the two classes is said to be *linearly separable*. In this case, it is always possible to rescale **w** and *b* so that

$$\min_{1 \leq i \leq k} y_i \left(\mathbf{w} \cdot \mathbf{x}_i + b\right) \geq 1$$

That is, the distance from the closest point to the hyperplane is $1/\|w\|$. Then (3) can be written as

$$y_i\left(\mathbf{w}\cdot\mathbf{x}_i+b\right)\geq 1 \qquad (4)$$

The hyperplane for which the distance to the closest point is maximal is called the *optimal separating hyperplane* (OSH) (Vapnik, 1995). As the distance to the closest point is $1/\|\mathbf{w}\|$, the OSH can be found by minimising $\|\mathbf{w}\|^2$ under constraint (4). The minimisation procedure uses Lagrange multipliers and Quadratic Programming (QP) optimisation methods. If $\lambda_i$, i = 1,….,k are the non-negative Lagrange multipliers associated with constraint (4), the optimisation problem becomes one of maximising (Osuna et.al. 1997):

$$L(\lambda)=\sum_i\lambda_i-\frac{1}{2}\sum_{i,j}\lambda_i\lambda_j y_i y_j(\mathbf{x}_i\cdot\mathbf{x}_j) \qquad (5)$$

under constraints $\lambda_i \geq 0$, i = 1, …..,k.

If $\lambda^a=(\lambda_1^a,......,\lambda_k^a)$ is an optimal solution of the maximisation problem (5) then the optimal separating hyperplane can be expressed as:

$$\mathbf{w}^a=\sum_i y_i\lambda_i^a\mathbf{x}_i \qquad (6)$$

The support vectors are the points for which $\lambda_i^a > 0$ when the equality in (4) holds.

If the data are not linearly separable, a slack variable $\xi_i$, i =1,……,k can be introduced with $\xi_i \geq 0$ (Cortes and Vapnik 1995) such that (4) can be written as

$$y_i\left(\mathbf{w}\cdot\mathbf{x}_i+b\right)-1+\xi_i\geq 0 \qquad (7)$$

and the solution to find a generalised OSH, also called a soft margin hyperplane, can be obtained using the conditions

$$\min_{\mathbf{w},b,\xi_1,....\xi_k}\left[\frac{1}{2}|\mathbf{w}|^2+C\sum_{i=1}^{k}\xi_i\right] \qquad (8)$$

$$y_i\left(\mathbf{w}\cdot\mathbf{x}_i+b\right)-1+\xi_i\geq 0 \qquad (9)$$

$$\xi_i\geq 0 \qquad i=1,\ \ldots\ldots k. \qquad (10)$$

The first term in (8) is same as in as in the linearly separable case, and controls the learning capacity, while the second term controls the number of misclassified points. The parameter C is chosen by the user. Larger values of C imply the assignment of a higher penalty to errors.

Where it is not possible to have a hyperplane defined by linear equations on the training data, the techniques described above for linearly separable data can be extended to allow for non-linear decision surfaces. A technique introduced by Boser et al. (1992) maps input data into a high dimensional feature space through some nonlinear mapping. The transformation to a higher dimensional space spreads the data out in a way that facilitates the finding of linear hyperplanes. After replacing **x** by its mapping in the feature space $\boldsymbol{\Phi}(\mathbf{x})$, equation (5) can be written as:

$$L(\boldsymbol{\lambda}) = \sum_i \lambda_i - \frac{1}{2}\sum_{i,j} \lambda_i \lambda_j y_i y_j \left(\boldsymbol{\Phi}(\mathbf{x_i}) \cdot \boldsymbol{\Phi}(\mathbf{x_j})\right) \qquad (11)$$

To reduce computational demands in feature space, it is convenient to introduce the concept of the *kernel function* K (Cristianini and Shawe-Taylor, 2000; Cortes and Vapnik 1995) such that:

$$K(\mathbf{x}_i, \mathbf{x}_j) = \boldsymbol{\Phi}(\mathbf{x}_i) \cdot \boldsymbol{\Phi}(\mathbf{x}_j) \qquad (12)$$

Then, to solve equation (11) only the kernel function is computed in place of computing $\boldsymbol{\Phi}(\mathbf{x})$, which could be computationally expensive. A number of kernel functions are used for support vector classifier. Details of some kernel functions and their parameters used with SVM classifiers are discussed by Vapnik (1995). SVM was initially designed for binary (two-class) problems. When dealing with several classes, an appropriate multi-class method is needed. A number of methods are suggested in literature to create multi-class classifiers using two-class methods (Hsu and Lin, 2002). In this study, a "one against one" approach (Knerr et al., 1990) (In this method, all possible two-class classifiers are evaluated from the training set of n classes, each classifier being trained on only two out of n classes. There would be a total of n (n-1)/2 classifiers. Applying each classifier to the vectors of the test data gives one vote to the winning class. The pixel is given the label of the class with most votes. To generate multi-class SVMs and a radial basis kernel function (defined as $e^{-\gamma|x-y|^2}$ ) was used.

## 3. Data

The first of the two study areas used in the work reported here are located near the town of Littleport in eastern England. The second is a wetland area of the La Mancha region of Spain. For the Littleport area, ETM+ data acquired on 19$^{th}$ June 2000 is used. The classification problem involves the identification of seven land cover types (wheat, potato, sugar beet, onion, peas, lettuce and beans) for the ETM+ data set. For the La Mancha study area, hyperspectral data acquired on 29$^{th}$ June 2000 by the DAIS 7915 airborne imaging spectrometer were available. Eight different land cover types (wheat, water body, dry salt lake, hydrophytic vegetation, vineyards, bare soil, pasture lands and built up area) were specified. The DAIS data show moderate to severe striping problems in the optical infrared region between bands 41 and 72. Initially, the first 72 bands in the wavelength range 0.4 μm to 2.5 μm were selected. All of these bands were examined visually to determine the severity of striping. Seven bands displaying very severe striping problems (bands 41 - 42 and 68 - 72) were removed from the data set. The striping in the remaining bands was removed by automatically enhancing the Fourier transform of each image (Cannon et al., 1983; Srinivasan et al., 1988). The input image is first divided into overlapping 128-by-128-pixel blocks. The Fourier transform of each block is calculated and the log-magnitudes of each FFT block are then averaged. The averaging process removes all frequency domain quantities except those which are present in each block; i.e., some sort of periodic interference. The average power spectrum is then used as a filter to adjust the FFT of the entire image. When an inverse Fourier transform is performed, the result is an image with periodic noise eliminated or significantly reduced.

## 4. Result and discussions

Random sampling was used to collect the training and test pixels for both ETM+ and DAIS data set. Total selected pixels were divided into two parts, one for training and one for testing the classifiers, so as to remove any possible bias resulting from the use of same set of pixels for both the testing and training phases. A total of 2700 training pixels and 2037 test pixels for ETM+ data and a total of 1600 training (200 pixels/class) and 3800 test pixels were used for DAIS data. Kappa values and overall classification accuracies are calculated for each of the classifiers used in this study with ETM+ data, while overall accuracy is calculated for the DAIS hyperspectral data. The Z statistic is also used to test the significance of apparent differences between the three classification algorithms, using ETM+ data. For this study, a standard back-propagation neural classifier was used. All user-defined parameters are set as recommended by Kavzoglu (2001).

Like neural network classifiers the performance of support vector classifier depends on a number of user-defined parameters which may influence the final classification accuracy. For this study a radial basis kernel with $\gamma$ (kernel specific parameter) value as two and C = 5000 is used for both data sets. The values of these parameters were chosen after a number of trials and the same parameters are used with DAIS data set. This study also suggests that, in comparison with the NN classifier, it is easier to fix the values of the user defined parameters for SVM.
As mentioned earlier, SVM involves in solving a quadratic programming problem with linear equality and inequality constraints which has only a global optimum. In comparison the presence of local minima is a significant problem in training the neural network classifiers.

Results obtained using ETM+ data suggests that support vector classifier perform well in comparison with neural and statistical classifier (Tables 1 and 2).

Table 1. Classification accuracies achieved with different classifiers.

| Classifier used | Accuracy (%) | Kappa value |
|---|---|---|
| Maximum likelihood | 82.9 | 0.80 |
| Neural network | 85.1 | 0.83 |
| Support vector | 87.9 | 0.87 |

Table 2. Calculated Z values for comparison between different classification systems with ETM+ data. Shaded values indicate significant improvements in the performance of first classifier at the 95% confidence level (Z critical value = 1.96).

| Classifiers | Z value |
|---|---|
| SVM vs. Neural network | 2.46 |
| SVM vs. maximum likelihood | 5.45 |

Further, the training time taken by support vector classifier is 0.30 minutes in comparison of 58 minutes by the NN classifier on a dual processor sun machine. Results suggest that support vector classifier performance is statistically significant in comparison with NN and ML classifiers. To study the behaviour of support vector classifier with DAIS Hyperspectral data a total of sixty five features (bands) was used, a total of 65 features (spectral bands) were available, as seven features with severe striping were discarded, as explained above. The initial number of features used was five, and the experiment was repeated with 10, 15, …, 65 features, giving a total of 13 experiments. Figure 1 suggests that, in comparison to the other classifiers, the performance of the support vector classifier is quite good with small training data set irrespective of the number of features used.

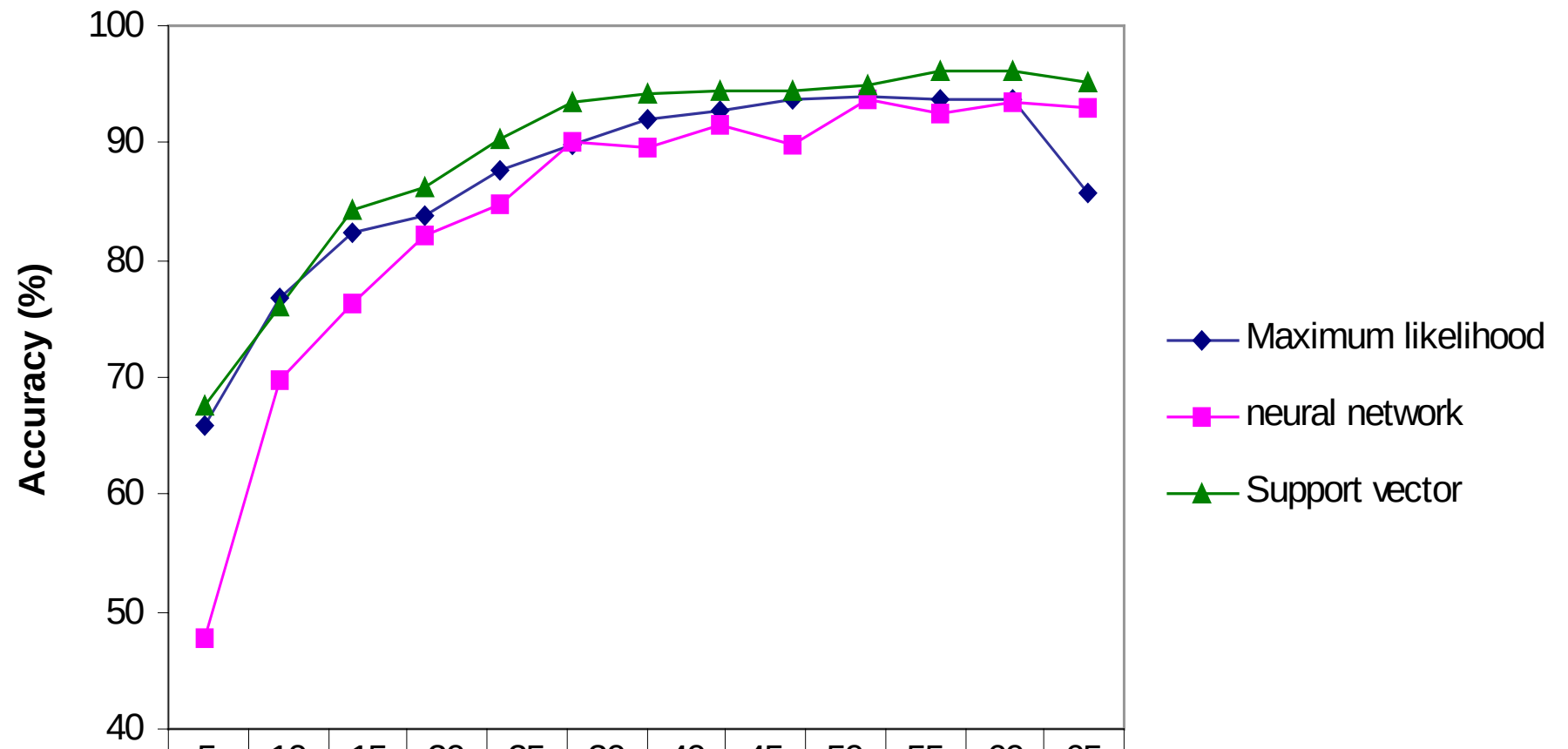

| | 5 | 10 | 15 | 20 | 25 | 30 | 40 | 45 | 50 | 55 | 60 | 65 |
|---|---|---|---|---|---|---|---|---|---|---|---|---|
| Maximum likelihood | 66 | 76.7 | 82.4 | 83.7 | 87.7 | 89.9 | 92.7 | 93.8 | 94 | 93.8 | 93.6 | 85.8 |
| neural network | 47.8 | 69.8 | 76.4 | 82.2 | 84.7 | 90.2 | 91.6 | 89.9 | 93.6 | 92.4 | 93.4 | 93 |
| Support vector | 67.6 | 76.1 | 84.3 | 86.2 | 90.3 | 93.4 | 94.5 | 94.5 | 95 | 96.1 | 96.1 | 95.1 |

Number of bands

Figure 1. Classification accuracies obtained with DAIS hyperspectral data using different classification algorithms. The training data set size is 200 pixels/class.

Results obtained from analysis of the hyperspectral data suggest that classification accuracy using SVM increases almost continuously as a function of the number of features, with the size of the training data set held constant, whereas the overall classification accuracies produced by the ML, DT and NN classifiers decline slightly once the number of bands exceeds 50 or so. Thus, suggesting that the 'Hughes phenomenon' (Hughes, 1968) of decreasing classifier performance as the dimensionality of the feature space increases beyond a threshold is not supported by the experiments using the support vector classifiers.

## 5. Conclusions

The objective of this study was to assess the utility of support vector classifiers for land cover classification using multi- and hyper-spectral data sets in comparison with most frequently used ML and NN classifiers. The results presented above suggest several conclusions. First the support vector classifier outperforms ML and NN classifiers in term of classification accuracy with both data sets. Several user-defined parameters affects the performance of the support vector classifier, but this study suggests that it is easier to find appropriate values for these parameters than it is for parameters defining the NN classifier. The level of classification accuracy achieved with the support vector classifier is better than both ML and NN classifiers when used with small number of training data.

## Acknowledgement

The DAIS data were collected and processed by DLR and were kindly made available by Prof. J. Gumuzzio of the Autonomous University of Madrid. Computing facilities were provided by the School of Geography, University of Nottingham.